\documentclass[runningheads]{llncs}
\usepackage[T1]{fontenc}
\usepackage[rgb,x11names]{xcolor}
\usepackage{amsmath}
\usepackage{amssymb}
\usepackage{subfig}
\usepackage{hyperref}
\hypersetup{colorlinks=true,citecolor=blue}
\usepackage{graphicx}
\usepackage{mathrsfs}
\usepackage{bbding}
\usepackage{rotating}
\usepackage{multirow}
\usepackage{marginnote}
\usepackage{overpic}
\usepackage{colortbl}
\usepackage[labelfont=bf]{caption}
\usepackage{bbding}
\usepackage{tabularx}
\usepackage{tikz}
\usepackage{float}
\usepackage{enumitem}
\usepackage{subcaption}
\usepackage{color}

\setlist[itemize]{label=\tiny$\bullet$}

\begin{document}
    \title{CAR-MFL: Cross-Modal Augmentation by Retrieval for Multimodal Federated Learning with Missing Modalities 
}

\author{Pranav Poudel\inst{2,5}\and
Prashant Shrestha \inst{2}$^{*}$ \and
Sanskar Amgain\inst{2}$^{*}$ \and
Yash Raj Shrestha\inst{3}  \and
Prashnna Gyawali \inst{4} \and 
Binod Bhattarai \inst{1}
}

%index{Poudel, Pranav}
%index{Shrestha, Prashant}
%index{Amgain, Sanskar}
%index{Shrestha, Yash Raj}
%index{Gyawali, Prashnna}
%index{Bhattarai, Binod}

\institute{University of Aberdeen, Aberdeen, UK\and NepAl Applied Mathematics and Informatics Institute for research, Nepal \and University of Lausanne, Switzerland \and West Virginia University, USA \and Fogsphere (Redev.AI), UK \\}

\authorrunning{P. Poudel {\it et al.}}

\renewcommand{\thefootnote}{*}
\footnotetext[1]{These authors made equal contributions.}
\renewcommand{\thefootnote}{\arabic{footnote}} 
\maketitle 
\begin{abstract}
Multimodal AI has demonstrated superior performance over unimodal approaches by leveraging diverse data sources for more comprehensive analysis. However, applying this effectiveness in healthcare is challenging due to the limited availability of public datasets. Federated learning presents an exciting solution, allowing the use of extensive databases from hospitals and health centers without centralizing sensitive data, thus maintaining privacy and security. Yet, research in multimodal federated learning, particularly in scenarios with missing modalities—a common issue in healthcare datasets—remains scarce, highlighting a critical area for future exploration.
 Toward this, we propose a novel method for multimodal federated learning with missing modalities. Our contribution lies in a novel cross-modal data augmentation by retrieval, leveraging the small publicly available dataset to fill the missing modalities in the clients. Our method learns the parameters in a federated manner, ensuring privacy protection and improving performance in multiple challenging multimodal benchmarks in the medical domain, surpassing several competitive baselines. Code Available: \url{https://github.com/bhattarailab/CAR-MFL}
 
\end{abstract}

\section{Introduction}
Multimodal AI is poised to revolutionize healthcare by integrating data from various modalities such as medical scans, omic data, and pathology reports \cite{acosta2022multimodal}. By synthesizing information across these different streams, such models offer a more holistic view of diseases, significantly enhancing diagnostic precision \cite{venugopalan2021multimodal,shrestha2023medical}. Yet, to learn such models, extensive training data is required, which poses a critical challenge, especially within healthcare, where privacy and security constraints severely limit data availability.

Federated learning offers a privacy-preserving approach to utilizing private datasets across medical centers, where learnings from private data can be aggregated to improve shared healthcare models without sharing raw data \cite{mcmahan2017communication,zhou2023fedcontrast,wang2023federated}. However, challenges arise in multimodal settings, as the modality distribution can be different across medical centers, due to clinical equipment unavailability, different acquisition procedures, or storage limitations 
% difficulty in storing all acquired data due to storage requirements 
\cite{thrasher2023multimodal}. 
Relevant approaches for handling missing modality scenarios in centralized settings often involve prompt-based \cite{seibold2022breaking,you2023cxr}, generative-based \cite{chen2023generative,lee2023unified} or dropout-based approaches \cite{lau2019unified,van2018learning}.
A notable work by Chen et al. \cite{chen2023towards} in the centralized domain employs cross-modal retrieval to impute the missing modality, where the complementary modality sample is retrieved by comparing features across modalities. However, this approach assumes shared representations between modalities, requiring additional contrastive objectives to enforce alignment.

Previous works that assume multimodal data to be available to all clients in a federated setting
\cite{sachin2023multimodal,qayyum2022collaborative,chen2024medical} disregard the challenges of missing modalities, where clients trained on a single modality may fail to properly capture the relationship between modalities, resulting in a divergence from multimodal clients and suboptimal federated training \cite{karimireddy2020scaffold,yu2023multimodal}.
A naive approach to dealing with missing modality is to simply replace the missing data with zeros \cite{zheng2023autofed,le2024cross}.  However, such a strategy can induce bias in the global model, causing it to potentially neglect representations learned from rarely present modalities.
Existing works considering missing modality in federated setting either assume access to a large-scale public multimodal dataset \cite{yu2023multimodal}, impractical in the context of medical data, or utilize class prototypes \cite{le2024cross} which may not be suitable for multi-label classification. Some approaches like sharing amplitude spectrum \cite{yan2023federated} are ill-defined for modalities like text and tabular data.
% \pranav{class prototype one impractical for multi-label classification problems}... 
% \sanskar{need to mention other approaches too, and how they are different from our works}. 
% \prashant{include computational overhead argument}

Towards this end, we propose a novel approach utilizing Cross-modal data Augmentation by Retrieval at the client level to address the missing modality in Multimodal Federated Learning. Our method, CAR-MFL leverages a small, publicly available multimodal dataset during training to retrieve the most likely absent modality samples for unimodal inputs. In doing so, we capitalize on the significant efforts made by the medical community to make certain datasets publicly available, while also scaling to include both multimodal and unimodal private datasets across medical centers.  To the best of our knowledge, this is the first work to augment a small-scale publicly available data set for Multimodal Federated Learning. While a recent work by Hao et al. ~\cite{hao2021towards} augments data at the client level for federated learning to mitigate the effects of data heterogeneity, it is limited to an unimodal setup. 
% Our method can seamlessly utilize multimodal synthetic data. 
Our focus, however, extends to maximizing the use of publicly available real multimodal data, often overlooked in federated learning scenarios. We summarize our key contributions below:
\begin{itemize}
 \item We introduce CAR-MFL, a novel multimodal federated learning framework employing retrieval-based cross-modal augmentation for missing modalities.
 \item We perform extensive evaluations, both quantitative and qualitative, on challenging benchmarks and compare with the competitive baselines, surpassing their performances.
\end{itemize}

\section{Method}
A typical multimodal federated setting consists of $K$ clients, each with its private dataset $D_k$ of $n_k$ samples.
The $i_{th}$ data sample in $D_k$ along with its corresponding label, $Y^{(i)}$ is denoted by the tuple 
$(\{X_m^{(i)}\}_{m=1}^{M_k}, Y^{(i)})$,
where $M_k$ represents the number of modalities in the $k^{th}$ client. Without loss of generality, we consider the presence of two modalities, image $I$ and text $T$. All clients share the same model architecture, comprising of modality-specific encoders $f_e$, a concatenation-based fusion module $\oplus$, and a classifier head $f_c$.
We thus represent the entire model $\boldsymbol{w}$ as a set $\{f_e^I, f_e^T, f_c\}$. 
The general global loss function to be optimized in multimodal federated setting is expressed as:
\begin{equation}
    \underset{w}{\operatorname{argmin}}~L(\boldsymbol{w}) =  \sum_{k = 1}^{K} \frac{|D_k|}{|D|}  \text{  } l_k(\boldsymbol{w}, D_k)
    \label{eq: global_optimization}
\end{equation}
\begin{equation}
    \text{where, }l_k(\boldsymbol{w}, D_k) = \frac{1}{|D_k|} \sum_{(X_I^{(i)}, X_T^{(i)}, Y^{(i)}) \in D_k} \hspace{-0.75cm} \mathcal{L}(f_c(f_e^I(X_I^{(i)}) \oplus f_e^T(X_T^{(i)})), Y^{(i)})
    \label{eq: local_optimization}
\end{equation}
Here, $D$ denotes the union of all client datasets, and $\mathcal{L}$ 
refers to the client-level local loss for the model $\{f_e^I, f_e^T, f_c\}$ on a data sample $(X_I^{(i)}, X_T^{(i)},Y^{(i)})$. 
A common approach to handling missing modality is zero-filling. Considering the case of text sample to be absent,. 
the local loss using zero-filling becomes $\mathcal{L}(f_c(f_e^I(X_I) \oplus 0), Y)$. Naively optimizing Eqn \ref{eq: global_optimization} in this scenario leads to modality biases where the resulting model tends to overlook the features of modality absent in majority of the clients, yielding suboptimal solutions.% \cite{?}.

\noindent \textbf{Cross-Modal Augmentation:}
\begin{figure}[ht]
    \centering
    \includegraphics[width=\textwidth]{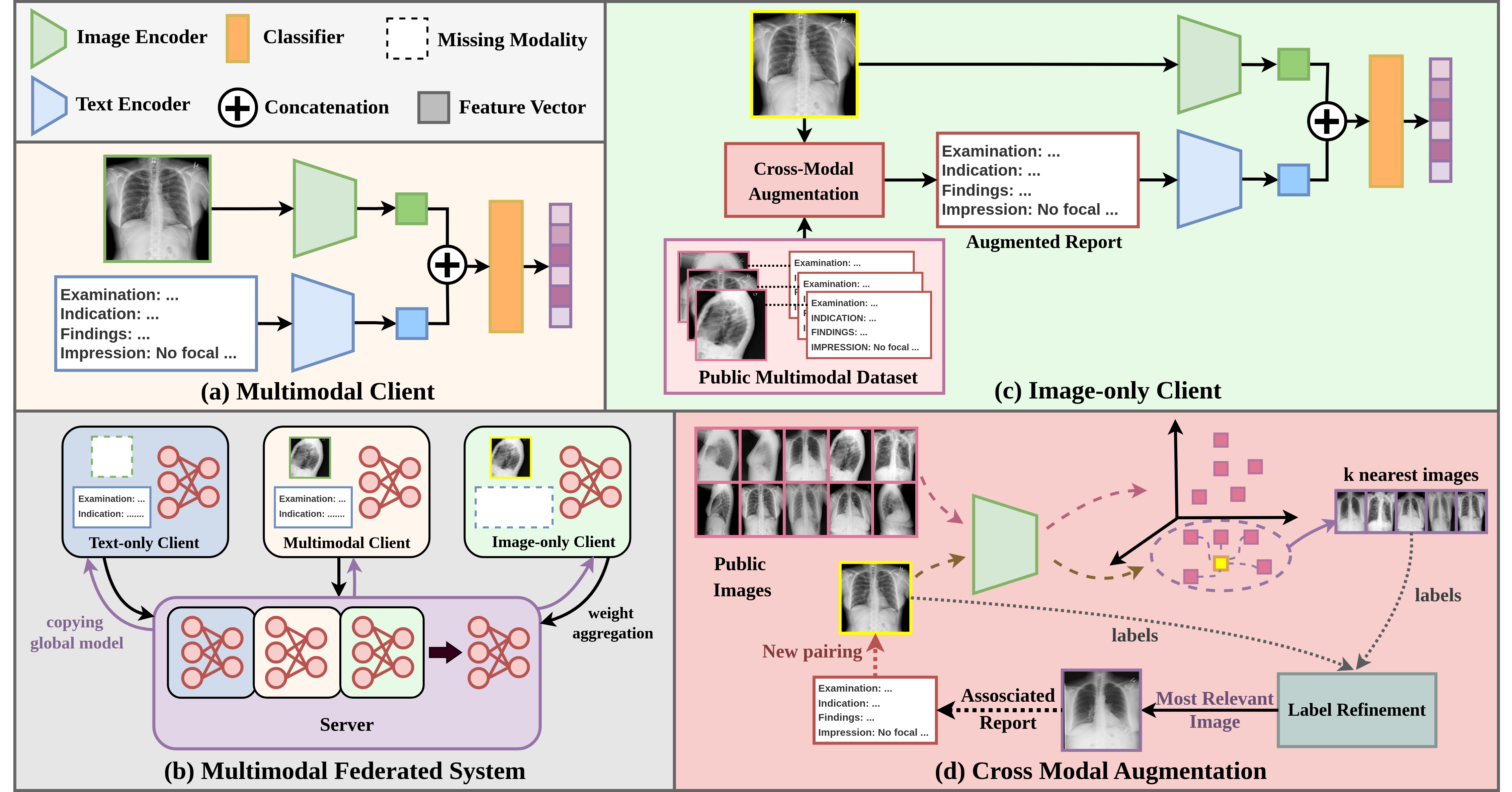}
    \caption{Illustration of \textbf{CAR-MFL}.
    \textbf{(a)} Multimodal client with access to multimodal data. \textbf{(b)} Multimodal federated system with missing modality. \textbf{(c)} Image client with only image samples; missing text modality is retrieved via our Cross-Modal Augmentation module. \textbf{(d)} Cross-Modal augmentation procedure for a query image (yellow): 
    Most relevant image from public data is retrieved based on distance in feature space and label similarity. Then, the associated text of the retrieved image is paired with the query image forming a paired input.
    }
    \label{fig:method}
\end{figure}
To address this challenge, we propose CAR-MFL, which utilizes cross-modal data augmentation for unimodal clients by utilizing a small publicly available multimodal dataset $D_p$ annotated and curated by the community.
This method creates new multimodal instances in unimodal clients by pairing the unimodal data with complementary modalities retrieved from the public dataset. This results in the following cases of augmented input:

\begin{equation}
    X = \begin{cases}
X_I, \textcolor{red}{T} & \text{if image-only client} \\
\textcolor{red}{I}~, X_T & \text{if text-only client} \\
X_I, X_T & \text{if multimodal client}
\end{cases}
    \label{eq:input_space_RCA}
\end{equation}
Here, $\textcolor{red}{I}$, $\textcolor{red}{T}$ are complementary modalities retrieved from the public dataset using cross-modal augmentation. Our overall method is as illustrated in Fig. \ref{fig:method}.

Commonly, complementary modalities are retrieved through direct cross-modal retrieval. This assumes aligned representation between modalities, necessitating additional cross-modal objectives for enforcement. Furthermore, such representations may not be optimal for classification tasks that could benefit from complementary information.
Our method first performs intra-modal retrieval and uses the associated modality of the retrieved public sample to impute the missing modality.
Consider an unimodal client with only image data (a similar process is applied to text-only clients). We begin with a pool of images $I_p = \{I_1, I_2,.., I_{N_p}\}$ from the public multimodal dataset $D_p$. At the start of each communication round, we select top-$k$ closest images, $I_{pk}$ from the public data pool for each image $X_I^c$ in the client dataset $D_c$. The selection is made based on the distance metric, $\delta$ between feature vectors of public images and local client images generated by the encoder of the global model, which is calculated as: 
\begin{equation}
\delta(X_I^{c(i)},I_i) = || f_e^I(X_I^{c(i)}) - f_e^I(I_i) ||^2
\label{eq:similarity score}
\end{equation}

Even intra-modal retrieval may yield semantically dissimilar samples at the beginning of the training procedure, owing to the lack of discriminative information in representation as the model is either pretrained on semantically different datasets or randomly initialized.
We thus propose to augment \textit{label refinement}, which further ranks the top-$k$ set $I_{pk}$ based on labels. 
This ranking prioritizes images with labels closely matching the query image, measured by Jaccard similarity between the label sets:
\begin{equation}
J(L_q, L_{pk}^j) = \frac{|L_q \cap L_{pk}^j|}{|L_q \cup L_{pk}^j|}
\label{eq:jaccard_metrics}
\end{equation}
where, $L_q$ denote the label set of $X_I^{c(i)}$ and $L_{pk}^j$ denote label sets of  $j^{th}$ image in $I_{pk}$; $I_j$.
We select the image, $I_r^{(i)}$, from $I_{pk}$ with the highest Jaccard similarity score using Eqn \ref{eq:jaccard_metrics}. This selected image has its associated text, $T_r^{(i)}$, which we pair with $X_I^{c(i)}$ to form complete multimodal data instances (Eqn. \ref{eq:DataAug}). This process is repeated for all the images in the local dataset, and a new augmented dataset, $D_c'$ is created, which is used for training the local model in unimodal client $c$:
\begin{equation}
     D_c' = \bigcup_{(X_I^{c(i)}, Y^{(i)}) \in D_c} (X_I^{c(i)}, \textcolor{red}{T_r^{(i)}}, Y^{(i)})
    \label{eq:DataAug}
\end{equation}

Using the augmented samples, the local loss function in Eqn. \ref{eq: local_optimization} becomes $\mathcal{L}(f_c(f_e^I(X_I^{(i)}) \oplus f_e^T(\textcolor{red}{T_r^{(i)}})), Y^{(i)})$. 
Retrieving these samples provides vital task-related information compared to the naive zero-filling approach. Consequently, the trained models learn to incorporate information from both modalities, mitigating the issue of modality bias.
In this way, we utilize the associated modality of the sample obtained by intra-model retrieval with label refinement to achieve cross-modal augmentation for missing modalities.

\noindent \textbf{Weight Re-adjustment:}
It is possible that the retrieved sample may not completely agree with the original label set $Y$ in every case. Using Eqn \ref{eq: global_optimization} directly could introduce noise and corrupt the global model. Hence, we perform aggregation weight re-adjustment by scaling down the existing weights ($\frac{|D_k|}{|D|}$) of the complementary modality encoders of unimodal clients with the hyper-parameter $\alpha$, followed by \textit{softmax} normalization to derive new re-adjusted weights. We use these adjusted weights only when updating the respective encoders, keeping the weights of the classification layer unchanged. We determine $\alpha=0.3$ through cross-validation (Refer to Supplementary Material).

\noindent \textbf{How do we address the Privacy Concerns?} It may raise questions for the reader that augmenting cross-modal data from publicly known identities may breach privacy protection. Our augmentation is a dynamic process that varies with the iterations (See Fig. \ref{fig:retrieval_qualitative}). It is evident that averaging the features from multiple identities obfuscates the identity and protects privacy through k-anonymity \cite{10.1007/11767831_15}. Moreover, the augmentations are done at the client level and pairing information is never shared with the other clients. 
\section{Experiments and Results}
\noindent \textbf{Datasets and Setups:}
We leverage three publicly available datasets: MIMIC-CXR \cite{johnson2019mimic}, NIH Open-I \cite{demner2016preparing}, and CheXpert \cite{irvin2019chexpert}, to create two experimental setups: Homogeneous and Heterogeneous.  
Both setups exclusively contain frontal views of chest X-rays and share a common public dataset comprising image-text pairs of 1000 patients sampled from MIMIC-CXR.  
The homogenous setup comprises 10 clients, each with training data for 810 patients sampled from the same dataset MIMIC-CXR. 
In contrast, the heterogeneous setup includes 8 image-only clients, each with data for 900 patients from CheXpert, and 2 multimodal clients, each with data for 1116 patients from NIH Open-I datasets.
This setup mirrors a real-world scenario where the data characteristics and label distribution across different categories of clients vary significantly. In both setups, the global model is validated and tested using common validation and test sets derived from the official splits of MIMIC-CXR. Additionally, each client can have multiple data instances. For brevity, we express client configurations in the format \textbf{I:T:M} where I and T denote the number of image-only and text-only clients respectively, and M indicates the number of multimodal clients in the setup.

\noindent \textbf{Implementation Details:}
We utilize pretrained Resnet50 \cite{he2016deep} and BERT-base \cite{devlin2018bert} architecture as image and text encoders respectively. Their outputs are reduced to 256-dimensional features and L2-normalized before fusion. The intra-modal retrieval is performed on these normalized features. We adopt a simple concatenation approach for multimodal fusion, followed by a linear layer for classification. The models are optimized using Adam \cite{kingma2014adam} with a learning rate of $1e^{-4}$ and are locally trained for $3$ epochs per communication round (30 total rounds). We retrieve the top 10 similar data points before subjecting them to label refinement to generate an augmented pair. We evaluate models using macro AUC (Average area under the Receiver Operating Characteristic Curve) on the multimodal test set. We use the labels generated by MedViLL \cite{moon2022multi} for Open-I. All reported values represent means from experiments conducted with 3 random seeds. All figures and studies, besides Table \ref{tab:perform} were conducted on the setting with 8 unimodal clients (8:0:2 and/or 0:8:2) within the homogeneous setup.

\noindent \textbf{Baselines:} We compared our method with three competitive baselines. First, we adapted FedAvg to handle multimodal data which we refer to as mFedAvg. Originally, FedAvg does not utilize publicly available datasets while training the model. To ensure a fair comparison, we train mFedAvg \emph{w} and 
\emph{w/o} the public data set, denoted as mFedAvg and mFedAvg\emph{P} respectively.  In both cases, we perform zero-fillings for the missing modalities. Both our method and mFedAvg\emph{P} train a separate model on the public data as a virtual client which is also used in aggregation. Finally, we compare our method to CreamFL~\cite{yu2023multimodal}, a state-of-the-art method for multimodal federated learning with missing modalities (ICLR'23).

\begin{table}[!t]
\centering
\setlength\tabcolsep{2pt}
\renewcommand{\arraystretch}{1.20}
\caption{AUC$\uparrow$ Performance in Homogeneous and Heterogeneous partitions. 
}
\begin{tabular}{ll|ccccccccc|c}
\hline
                                                       & Partitions      & \multicolumn{3}{c}{}                                                  & \multicolumn{3}{c}{Homogeneous}                                              & \multicolumn{3}{c|}{}                            & Hetero \\ \hline
                                                       & (I:T:M)         & 4:0:6          & 0:4:6          & \multicolumn{1}{c|}{2:2:6}          & 6:0:4          & 0:6:4          & \multicolumn{1}{c|}{3:3:4}          & 8:0:2          & 0:8:2          & 4:4:2          & 8:0:2  \\ \hline
\multirow{4}{*}{\begin{sideways}Methods\end{sideways}} & mFedAvg         & 86.64          & 87.95          & \multicolumn{1}{c|}{87.4}           & 82.99          & 87.01          & \multicolumn{1}{c|}{84.48}          & 79.69          & 86.32          & 80.47          & 72.76  \\
                                                       & mFedAvg\emph{P} & 87.77          & 88.81          & \multicolumn{1}{c|}{88.69}          & 85.49          & 87.64          & \multicolumn{1}{c|}{85.78}          & 81.95          & 87.13          & 83.46          & 82.52  \\
                                                       & CreamFL         & 78.42          & 75.39          & \multicolumn{1}{c|}{75.79}          & 78.47          & 75.47          & \multicolumn{1}{c|}{75.65}          & 78.36          & 75.28          & 75.49          & 78.22  \\
                                                       & CAR-MFL         & \textbf{89.78} & \textbf{90.06} & \multicolumn{1}{c|}{\textbf{89.94}} & \textbf{88.22} & \textbf{90.12} & \multicolumn{1}{c|}{\textbf{89.62}} & \textbf{87.31} & \textbf{89.22} & \textbf{88.93} & \textbf{86.14}  \\ \hline
\end{tabular}
\label{tab:perform}
\end{table}

\begin{figure}[ht]
    \centering
    \includegraphics[scale=0.39]{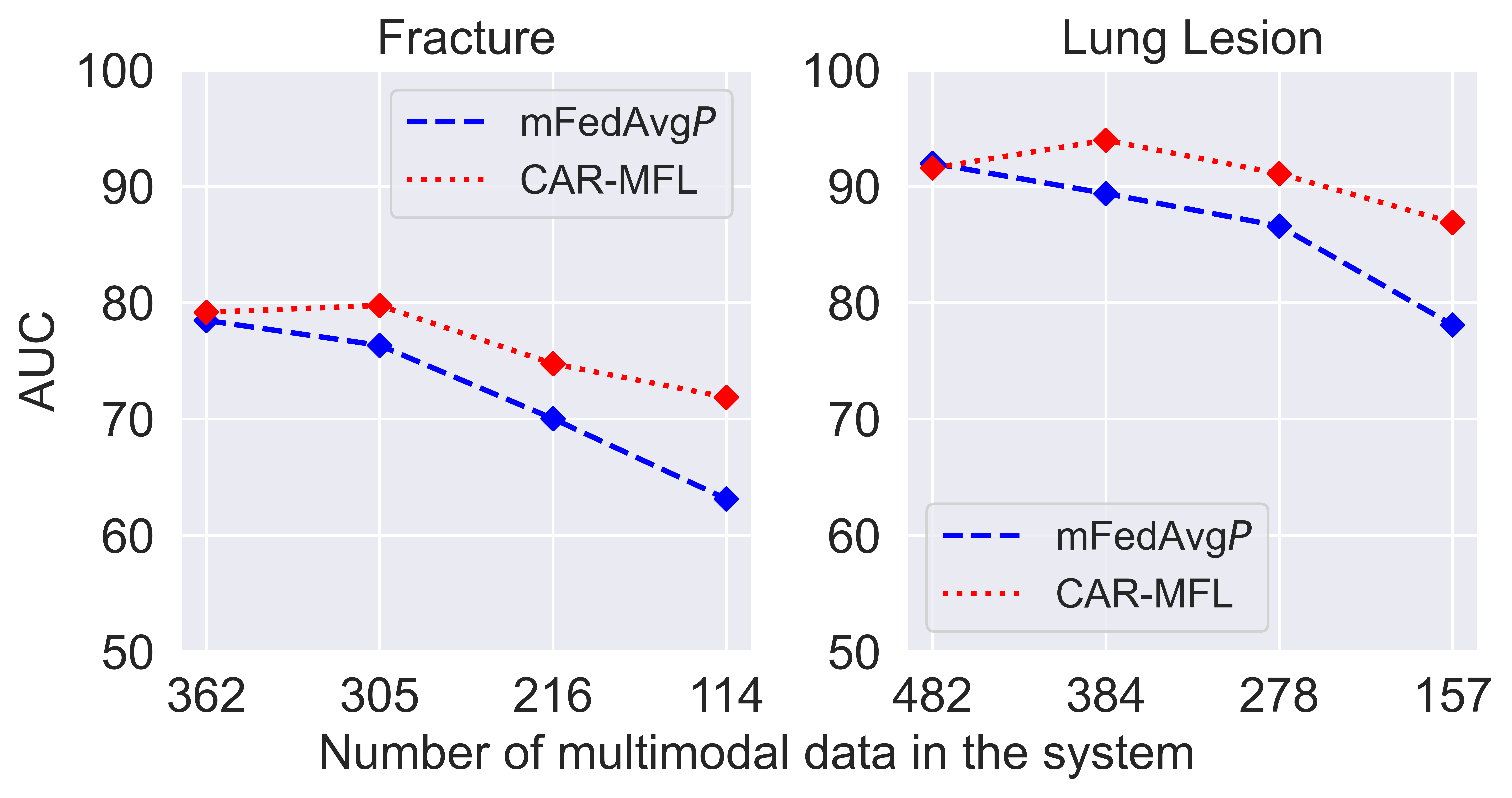}
    \caption{Comparison between mFedAvg\emph{P} and CAR-MFL on rare pathologies.}
    \label{fig:clinical_relevancy}
\end{figure}

\subsection{Quantitative Results}
 Table \ref{tab:perform} shows the AUC of all baselines and our method across various client configurations in both setups. Our method, CAR-MFL significantly outperforms all baselines at all settings of modality distribution in the homogeneous as well as challenging heterogeneous partition demonstrating its applicability in real-world settings.
In the homogeneous setup, CAR-MFL, with just 2 multimodal clients, surpasses the mFedAvg\emph{P} model employing 6 multimodal clients. 
Furthermore, CAR-MFL outperforms mFedAvg\emph{P} in the heterogeneous setup by 4\%.
The poor performance of CreamFL is attributed to their assumption of aligned representations between modalities and their imposition through data-intensive inter-modal loss.
Furthermore, their utilization of 50,000 public data samples is in stark contrast to our setups which utilize just 2701 paired samples as the public data. 
For additional results and findings, please refer to Supplementary section.

\noindent \textbf{Clinical Relevancy}: 
To investigate the clinical applicability of our proposed framework, we analyzed its performance with a focus on rare pathological classes. This evaluation specifically targeted how the absence of multimodal data influences the diagnostic accuracy of such rare conditions. We focused on two rare categories: "Fracture" and "Lung Lesion", with less than 1.5\% of instances out of 43,802 total instances, progressively reducing the share of multimodal clients to simulate a transition towards a predominantly unimodal dataset within the system. As shown in Fig. \ref{fig:clinical_relevancy}, despite an anticipated decrease in the effectiveness of the CAR-MFL approach under these conditions, it exhibited robustness compared to the mFedAvg\emph{P} framework, which demonstrated a significant drop in performance when moving from a multimodal to a unimodal client setup. This stark difference in performance, particularly noted in the handling of these rare classes, underscores the superior clinical relevance of our model.
\begin{figure}[t]
    \centering
    \includegraphics[scale=0.37]{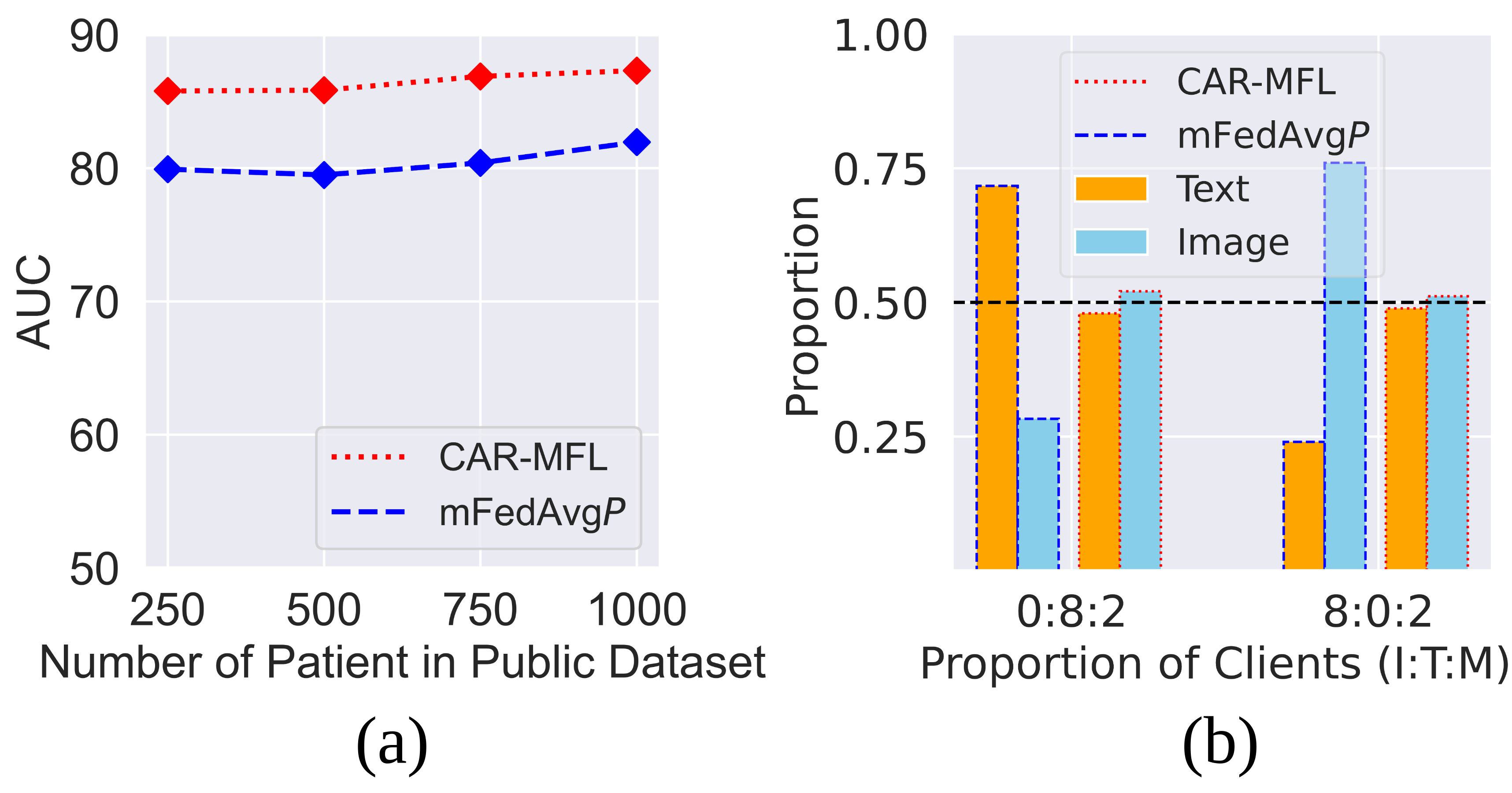}
    \caption{\textbf{(a)} Study of model AUC on varying public data size. \textbf{(b)} Distribution of weight values of classification layer across image and text features.}
    \label{fig:public_data_weight_distributions}
\end{figure}

\noindent \textbf{Study on Public Data Size}: In Fig. \ref{fig:public_data_weight_distributions} a, We evaluate the effectiveness of our approach at various levels of public training data keeping the test set unchanged. CAR-MFL demonstrates substantial performance improvement compared to mFedAvg\emph{P} and robustness to decreasing public data. Remarkably, even with data from just 250 patients, CAR-MFL outperforms mFedAvg\emph{P} by 6\%.

\subsection{Qualitative Results}
\noindent \textbf{Modality Bias}:
We examine the distribution of classification layer weights to assess the effectiveness of our approach in mitigating modality bias. We calculate the distribution by summing the absolute values of the weights for image and text features separately, then normalize them by the total weights. As depicted in Fig. \ref{fig:public_data_weight_distributions} b, mFedAvg\emph{P} with zero-filling, exhibits a distinct bias, with most weight values allocated to the modality present in the majority of clients. In contrast, CAR-MFL shows a near-equal weight distribution for both modalities.

\begin{figure}[t]
    \centering
    \includegraphics[scale=0.5]{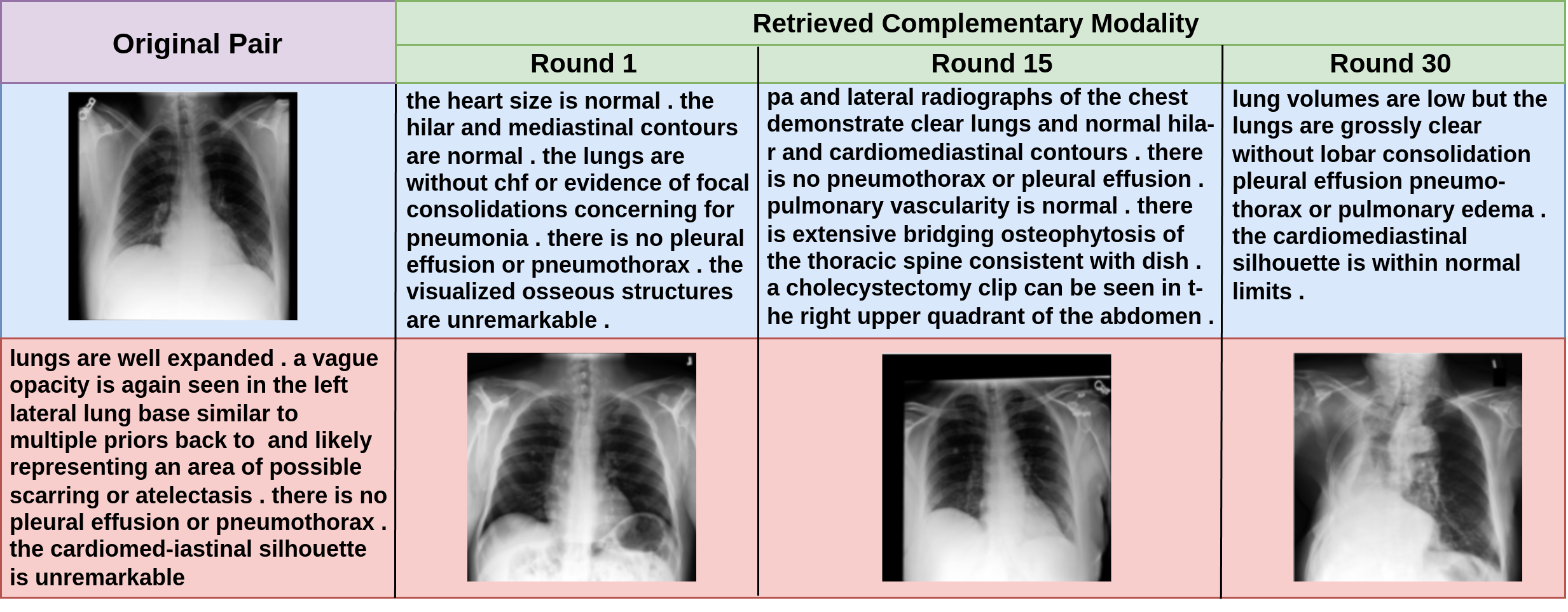}
    \caption{Qualitative Analysis of retrieved samples across different training rounds. Column 1 contains a paired image text sample. The first row displays retrieved text reports from the public dataset when the text modality is missing. The second row displays retrieved images when the image modality is missing.}
\label{fig:retrieval_qualitative}
\end{figure}

\noindent\textbf{Dynamic Augmentation:} We additionally evaluate retrieval samples obtained by our cross-modal augmentation module at various stages of training in Fig. \ref{fig:retrieval_qualitative}. We observed that, on average, 
an image sample in the unimodal image client is paired with 16 unique text samples, while a text sample in a text client is paired with 9 unique image samples. This evaluation was done on the 8 unimodal client settings (0:8:2 and 8:0:2) over 30 training rounds.

\section{Conclusion}
We presented a retrieval-based cross-modal augmentation approach for multimodal federated learning with missing modalities. This approach imputes the absent modality samples in unimodal clients utilizing a small public multimodal dataset. Our simple yet effective method surpasses previous federated learning frameworks, including the recently proposed state-of-the-art method, CreamFL, while requiring significantly fewer public data samples. Moreover, our approach does not assume a common representation space for different modalities and adds no extra computational overhead to the training procedure, ensuring its broad applicability across various multimodal tasks.

\begin{credits}
\subsubsection{\ackname} This project is supported by the University of Aberdeen Startup grant CF10834-10.
% \subsubsection{\discintname}
% The authors have no competing interests to declare that are relevant to the content of this article.
\end{credits}
\bibliographystyle{splncs04}
\bibliography{Paper-0391}

\begin{thebibliography}{10}
\providecommand{\url}[1]{\texttt{#1}}
\providecommand{\urlprefix}{URL }
\providecommand{\doi}[1]{https://doi.org/#1}

\bibitem{acosta2022multimodal}
Acosta, J.N., Falcone, G.J., Rajpurkar, P., Topol, E.J.: Multimodal biomedical ai. Nature Medicine  \textbf{28}(9),  1773--1784 (2022)

\bibitem{chen2024medical}
Chen, J., Pan, R.: Medical report generation based on multimodal federated learning. Computerized Medical Imaging and Graphics p. 102342 (2024)

\bibitem{chen2023generative}
Chen, Y., Liu, C., Huang, W., Cheng, S., Arcucci, R., Xiong, Z.: Generative text-guided 3d vision-language pretraining for unified medical image segmentation. arXiv preprint arXiv:2306.04811  (2023)

\bibitem{chen2023towards}
Chen, Z., Diao, S., Wang, B., Li, G., Wan, X.: Towards unifying medical vision-and-language pre-training via soft prompts. arXiv preprint arXiv:2302.08958  (2023)

\bibitem{demner2016preparing}
Demner-Fushman, D., Kohli, M.D., Rosenman, M.B., Shooshan, S.E., Rodriguez, L., Antani, S., Thoma, G.R., McDonald, C.J.: Preparing a collection of radiology examinations for distribution and retrieval. Journal of the American Medical Informatics Association  \textbf{23}(2),  304--310 (2016)

\bibitem{devlin2018bert}
Devlin, J., Chang, M.W., Lee, K., Toutanova, K.: Bert: Pre-training of deep bidirectional transformers for language understanding. arXiv preprint arXiv:1810.04805  (2018)

\bibitem{10.1007/11767831_15}
Gross, R., Airoldi, E., Malin, B., Sweeney, L.: Integrating utility into face de-identification. In: Danezis, G., Martin, D. (eds.) Privacy Enhancing Technologies. pp. 227--242. Springer Berlin Heidelberg, Berlin, Heidelberg (2006)

\bibitem{hao2021towards}
Hao, W., El-Khamy, M., Lee, J., Zhang, J., Liang, K.J., Chen, C., Duke, L.C.: Towards fair federated learning with zero-shot data augmentation. In: Proceedings of the IEEE/CVF Conference on Computer Vision and Pattern Recognition. pp. 3310--3319 (2021)

\bibitem{he2016deep}
He, K., Zhang, X., Ren, S., Sun, J.: Deep residual learning for image recognition. In: Proceedings of the IEEE conference on computer vision and pattern recognition. pp. 770--778 (2016)

\bibitem{irvin2019chexpert}
Irvin, J., Rajpurkar, P., Ko, M., Yu, Y., Ciurea-Ilcus, S., Chute, C., Marklund, H., Haghgoo, B., Ball, R., Shpanskaya, K., et~al.: Chexpert: A large chest radiograph dataset with uncertainty labels and expert comparison. In: Proceedings of the AAAI conference on artificial intelligence. vol.~33, pp. 590--597 (2019)

\bibitem{johnson2019mimic}
Johnson, A.E., Pollard, T.J., Berkowitz, S.J., Greenbaum, N.R., Lungren, M.P., Deng, C.y., Mark, R.G., Horng, S.: Mimic-cxr, a de-identified publicly available database of chest radiographs with free-text reports. Scientific data  \textbf{6}(1), ~317 (2019)

\bibitem{karimireddy2020scaffold}
Karimireddy, S.P., Kale, S., Mohri, M., Reddi, S., Stich, S., Suresh, A.T.: Scaffold: Stochastic controlled averaging for federated learning. In: International conference on machine learning. pp. 5132--5143. PMLR (2020)

\bibitem{kingma2014adam}
Kingma, D.P., Ba, J.: Adam: A method for stochastic optimization. arXiv preprint arXiv:1412.6980  (2014)

\bibitem{lau2019unified}
Lau, K., Adler, J., Sj{\"o}lund, J.: A unified representation network for segmentation with missing modalities. arXiv preprint arXiv:1908.06683  (2019)

\bibitem{le2024cross}
Le, H.Q., Thwal, C.M., Qiao, Y., Tun, Y.L., Nguyen, M.N., Hong, C.S.: Cross-modal prototype based multimodal federated learning under severely missing modality. arXiv preprint arXiv:2401.13898  (2024)

\bibitem{lee2023unified}
Lee, H., Kim, W., Kim, J.H., Kim, T., Kim, J., Sunwoo, L., Choi, E.: Unified chest x-ray and radiology report generation model with multi-view chest x-rays. arXiv preprint arXiv:2302.12172  (2023)

\bibitem{mcmahan2017communication}
McMahan, B., Moore, E., Ramage, D., Hampson, S., y~Arcas, B.A.: Communication-efficient learning of deep networks from decentralized data. In: Artificial intelligence and statistics. pp. 1273--1282. PMLR (2017)

\bibitem{moon2022multi}
Moon, J.H., Lee, H., Shin, W., Kim, Y.H., Choi, E.: Multi-modal understanding and generation for medical images and text via vision-language pre-training. IEEE Journal of Biomedical and Health Informatics  \textbf{26}(12),  6070--6080 (2022)

\bibitem{qayyum2022collaborative}
Qayyum, A., Ahmad, K., Ahsan, M.A., Al-Fuqaha, A., Qadir, J.: Collaborative federated learning for healthcare: Multi-modal covid-19 diagnosis at the edge. IEEE Open Journal of the Computer Society  \textbf{3},  172--184 (2022)

\bibitem{sachin2023multimodal}
Sachin, D., Annappa, B., Ambasange, S., Tony, A.E.: A multimodal contrastive federated learning for digital healthcare. SN Computer Science  \textbf{4}(5), ~674 (2023)

\bibitem{seibold2022breaking}
Seibold, C., Rei{\ss}, S., Sarfraz, M.S., Stiefelhagen, R., Kleesiek, J.: Breaking with fixed set pathology recognition through report-guided contrastive training. In: International Conference on Medical Image Computing and Computer-Assisted Intervention. pp. 690--700. Springer (2022)

\bibitem{shrestha2023medical}
Shrestha, P., Amgain, S., Khanal, B., Linte, C.A., Bhattarai, B.: Medical vision language pretraining: A survey. arXiv preprint arXiv:2312.06224  (2023)

\bibitem{thrasher2023multimodal}
Thrasher, J., Devkota, A., Siwakotai, P., Chivukula, R., Poudel, P., Hu, C., Bhattarai, B., Gyawali, P.: Multimodal federated learning in healthcare: a review. arXiv preprint arXiv:2310.09650  (2023)

\bibitem{van2018learning}
van Tulder, G., de~Bruijne, M.: Learning cross-modality representations from multi-modal images. IEEE transactions on medical imaging  \textbf{38}(2),  638--648 (2018)

\bibitem{venugopalan2021multimodal}
Venugopalan, J., Tong, L., Hassanzadeh, H.R., Wang, M.D.: Multimodal deep learning models for early detection of alzheimer’s disease stage. Scientific reports  \textbf{11}(1), ~3254 (2021)

\bibitem{wang2023federated}
Wang, M., Wang, L., Xu, X., Zou, K., Qian, Y., Goh, R.S.M., Liu, Y., Fu, H.: Federated uncertainty-aware aggregation for fundus diabetic retinopathy staging. arXiv preprint arXiv:2303.13033  (2023)

\bibitem{yan2023federated}
Yan, Y., Feng, C.M., Li, Y., Goh, R.S.M., Zhu, L.: Federated pseudo modality generation for incomplete multi-modal mri reconstruction. arXiv preprint arXiv:2308.10910  (2023)

\bibitem{you2023cxr}
You, K., Gu, J., Ham, J., Park, B., Kim, J., Hong, E.K., Baek, W., Roh, B.: Cxr-clip: Toward large scale chest x-ray language-image pre-training. In: International Conference on Medical Image Computing and Computer-Assisted Intervention. pp. 101--111. Springer (2023)

\bibitem{yu2023multimodal}
Yu, Q., Liu, Y., Wang, Y., Xu, K., Liu, J.: Multimodal federated learning via contrastive representation ensemble. arXiv preprint arXiv:2302.08888  (2023)

\bibitem{zheng2023autofed}
Zheng, T., Li, A., Chen, Z., Wang, H., Luo, J.: Autofed: Heterogeneity-aware federated multimodal learning for robust autonomous driving. arXiv preprint arXiv:2302.08646  (2023)

\bibitem{zhou2023fedcontrast}
Zhou, Q., Zheng, G.: Fedcontrast-gpa: Heterogeneous federated optimization via local contrastive learning and global process-aware aggregation. In: International Conference on Medical Image Computing and Computer-Assisted Intervention. pp. 660--670. Springer (2023)

\end{thebibliography}
\newpage
\section{Supplementary Materials}
\begin{figure}[ht]
    \centering
\includegraphics[width=\textwidth]{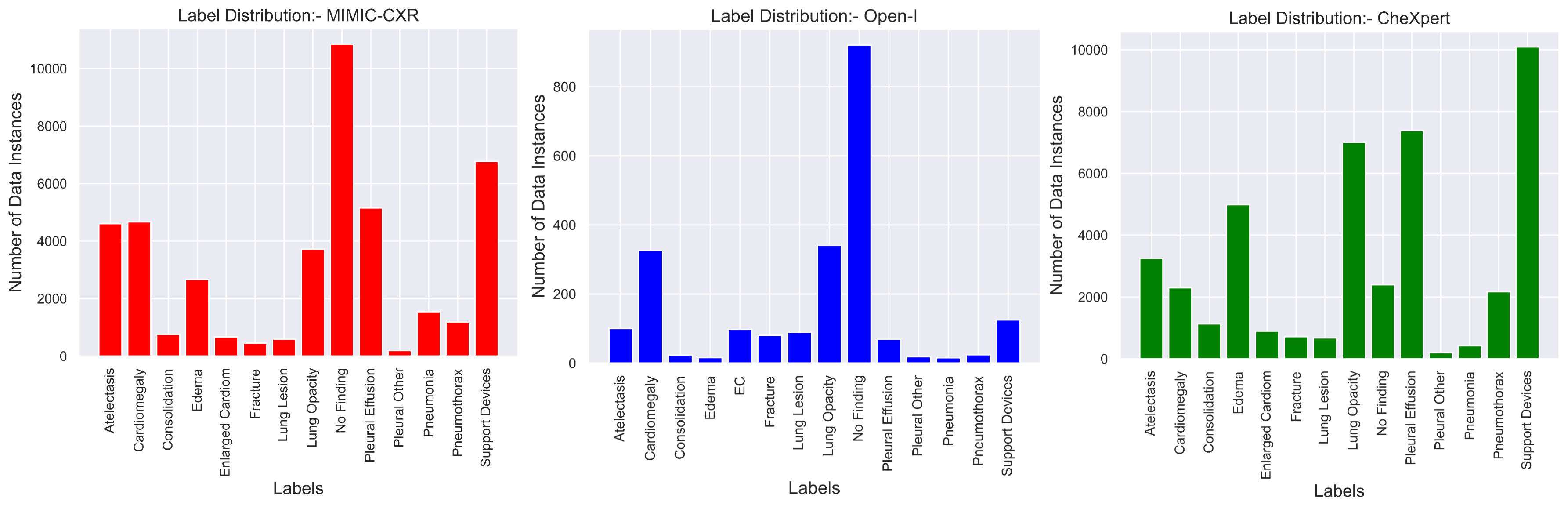}
    \caption{Label distribution of different datasets in our setup.}
\label{fig:label-distro}
    \vspace{-30pt}
\end{figure}
% \vspace{-1.25cm}
\begin{table}[!ht]
\centering
\setlength\tabcolsep{2pt}
\renewcommand{\arraystretch}{1.2}
\caption{Comparison with upper and lower bound values in homogeneous setup: Lowerbound refers to a model trained only with public data. The Upperbound refers to when all data is stored in a central server including public data. mFedAvg\textit{P}-NM refers to mFedAvgP in the case where there are no missing modalities in clients. Both mFedAvg\textit{P} and CAR-MFL values are of the extreme setting of 8 image-only clients (8:0:2).}
\label{tab:bound}
\begin{tabular}{c|ccccc}
\hline
% & & \textbf{Methods} &  \\
& \textbf{Upperbound} & \textbf{Lowerbound} & \textbf{mFedAvg\textit{P}-NM} & \textbf{mFedAvg\textit{P}} & \textbf{CAR-MFL}\\ \hline
\textbf{AUC}    & 91.67      & 83.11      & 90.17    & 81.95 & 87.31\\ \hline
\end{tabular}
\end{table}
\begin{table}[!ht]
\centering
\setlength\tabcolsep{2pt}
\renewcommand{\arraystretch}{1.2}
\caption{No. of data samples at various patient counts in public data.}
\begin{tabular}{l|llll}
\hline
\textbf{No. of Patients}     & 1000 & 750  & 500  & 250 \\ \hline
\textbf{No. of Data Samples} & 2701 & 1888 & 1210 & 602 \\ \hline
\end{tabular}
\end{table}
\begin{table}[!ht]
\setlength\tabcolsep{3pt}
\renewcommand{\arraystretch}{1.20}
\centering
\caption{No. of data samples across clients in \textit{homogeneous} and \textit{heterogeneous} setups. In a heterogeneous setup, clients 8 and 9 contain multimodal data from NIH Open-I, and the rest are image-only clients with images from CheXpert.}
\label{tab:data-distribution}
\begin{tabular}{l|cccccccccc}
\hline
\textbf{Client ID}    & 0    & 1    & 2    & 3    & 4    & 5    & 6    & 7    & 8    & 9    \\ \hline
\textbf{Homogeneous}  & 2343 & 2123 & 2171 & 2107 & 2195 & 2127 & 2164 & 2188 & 2528 & 2086 \\
\textbf{Heterogeneous} & 2245 & 2154 & 2113 & 2359 & 2133 & 2003 & 2189 & 2205 & 1116 & 1116 \\ \hline
\end{tabular}
\end{table}
\begin{table}[!ht]
\centering
\setlength\tabcolsep{2pt}
\renewcommand{\arraystretch}{1.6}
\caption{Validation AUC across the various $\alpha$ for homogeneous  4:0:6 setting.}
\label{tab:alpha}
\begin{tabular}{c|cccccc}
\hline
$\boldmath\alpha$ & 1     & 0.5   & 0.4   & \textbf{0.3}   & 0.2   & 0     \\ \hline
\textbf{AUC}      & 92.16 & 91.98 & 92.19 & \textbf{92.21} & 91.43 & 91.95 \\ \hline
\end{tabular}
% \vspace{-260pt}
\end{table}
\begin{figure}[ht]
    \centering
\includegraphics[scale=0.45]{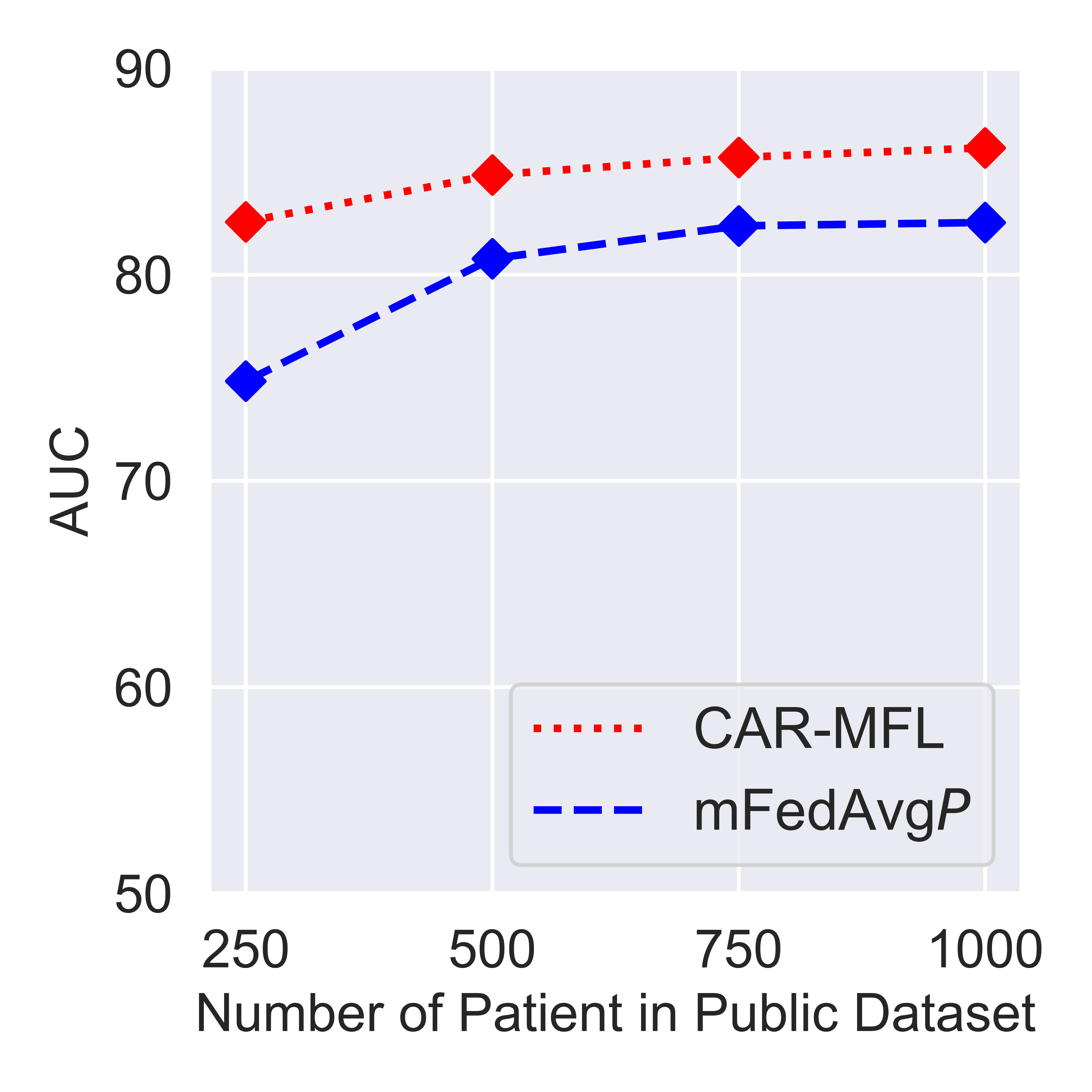}
    \caption{Model AUC on varying public data size for \textit{heterogeneous} setup.}
\label{fig:hetero_public}
\end{figure}
\begin{figure}[ht]
    \centering
    \includegraphics[width=\textwidth]{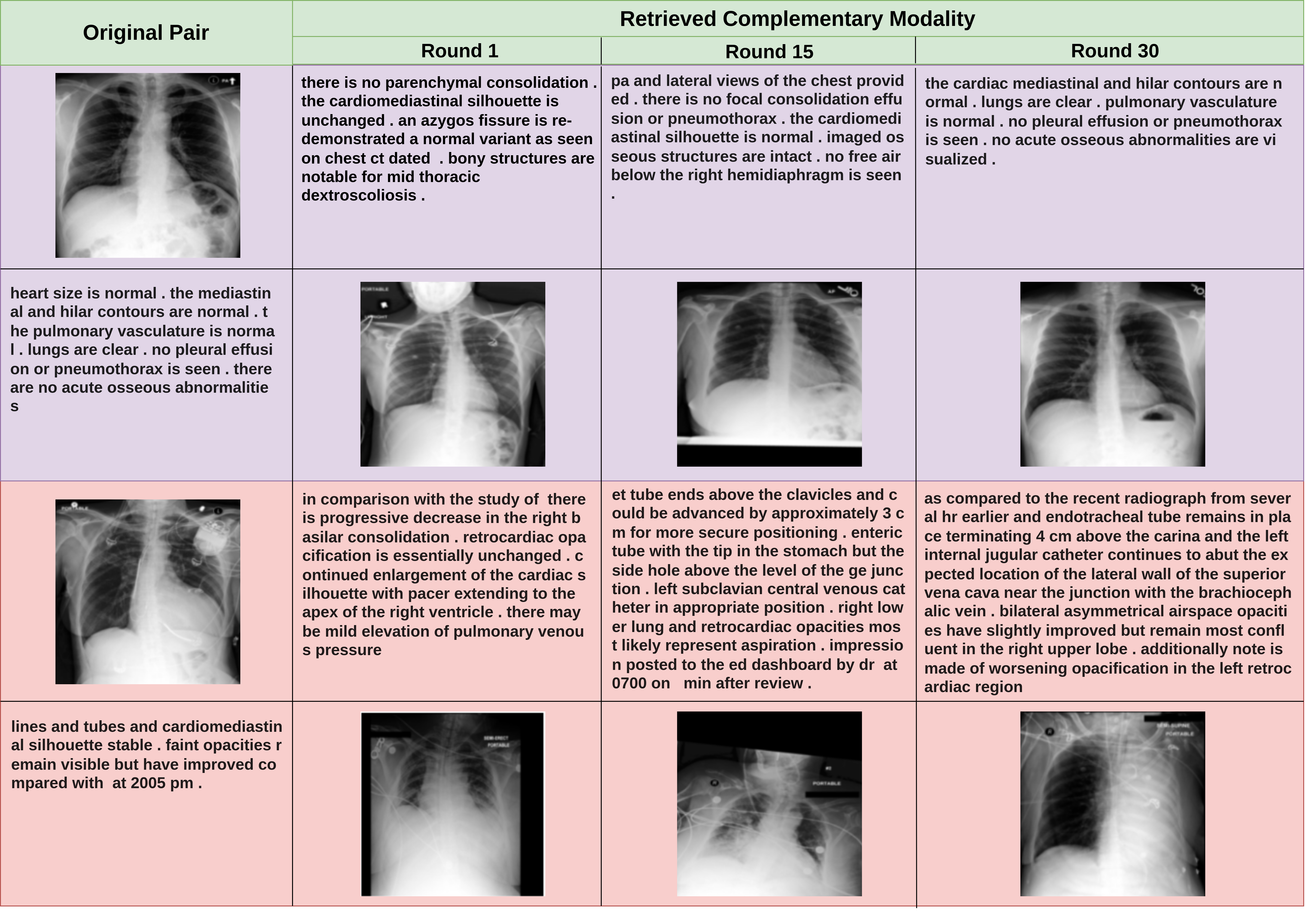}
    \caption{Qualitative Analysis of retrieved samples across different training rounds. Column 1 contains two paired image text samples. Row 1 and 3 display text reports from the public dataset when the text modality is missing. Row 2 and 4 display retrieved images when the image modality is missing.}
    % \caption{Qualitative Analysis of retrieved samples. The top row displays retrieved text reports from a public dataset across communication rounds and The bottom row shows the retrieved image data. Both rows originate from the same data pairs. The top row represents instances where the client acts as an image client, while the bottom row shows interactions where the client acts as a text client. The first column presents the original image-text pairs}
\label{fig:retrieval_qualitative_append}
\end{figure}

\end{document}